\documentclass{article}
\usepackage{xcolor}
\usepackage{spconf,amsmath,graphicx}
\usepackage{amsfonts}
\usepackage{mathrsfs}
\usepackage{multirow}
\usepackage{multicol}
\usepackage{booktabs}
\usepackage[colorlinks,
linkcolor=blue,
anchorcolor=black,
citecolor=black]{hyperref}
\usepackage{url}

\title{Towards Intelligent Design: A Self-driven Framework for Collocated Clothing Synthesis Leveraging Fashion Styles and Textures}
\name{Minglong Dong$^{1 \star}$, Dongliang Zhou$^{1 \star}$, Jianghong Ma$^{1, 2}$, and Haijun Zhang$^{1}$\thanks{\fontsize{7.4pt}{7.4pt}\selectfont Asterisk indicates that the first two authors contributed equally to this work. The order of authorship was determined by a coin toss. This work was supported in part by the National Natural Science Foundation of China under Grant 62202122, and 62073272, the Guangdong Basic and Applied Basic Research Foundation under Grant 2021B1515020088, the Shenzhen Science and Technology Program under Grant JCYJ20210324131203009, and the Harbin Institute of Technology (Shenzhen) (HITSZ)-J$\&$A Joint Laboratory of Digital Design and Intelligent Fabrication under Grant HITSZ-J$\&$A- 021A01. Corresponding author: Haijun Zhang, e-mail: hjzhang@hit.edu.cn.}}
\address{$^{1}$ Harbin Institute of Technology, Shenzhen, China \\ $^{2}$ City University of Hong Kong, Hong Kong, China}

\begin{document}
%
\maketitle
\begin{abstract}
Collocated clothing synthesis (CCS) has emerged as a pivotal topic in fashion technology, primarily concerned with the generation of a clothing item that harmoniously matches a given item. However, previous investigations have relied on using paired outfits, such as a pair of matching upper and lower clothing, to train a generative model for achieving this task. This reliance on the expertise of fashion professionals in the construction of such paired outfits has engendered a laborious and time-intensive process. In this paper, we introduce a new self-driven framework, named style- and texture-guided generative network (ST-Net), to synthesize collocated clothing without the necessity for paired outfits, leveraging self-supervised learning. ST-Net is designed to extrapolate fashion compatibility rules from the style and texture attributes of clothing, using a generative adversarial network. To facilitate the training and evaluation of our model, we have constructed a large-scale dataset specifically tailored for unsupervised CCS. Extensive experiments substantiate that our proposed method outperforms the state-of-the-art baselines in terms of both visual authenticity and fashion compatibility.
\end{abstract}
\vspace{-0.22cm}
\section{Introduction}
\vspace{-0.22cm}
\label{sec:intro}
In the era of artificial intelligence-generated content (AIGC), fashion design has gained unprecedented prominence, driven by the pervasive utilization of generative adversarial networks (GANs) \cite{bodur2023joint,karras2019style,huang2023cpd} and diffusion models \cite{rombach2022high}. A specific task within this realm, referred to as collocated clothing synthesis (CCS) \cite{liu2019collocating,yu2019personalized,zhou2022coutfitgan,zhou2023fcboost} has garnered considerable scholarly attention. CCS focuses on synthesizing a complementary item of clothing to match the given item. For instance, given an item of upper clothing, it can generate a collocated item of lower clothing. However, previous endeavors in CCS rely on constructed datasets of matching outfits, a construction process that inherently demands the judicious involvement of domain experts, thus incurring substantial temporal and labor costs. Simultaneously, the vast reservoir of unpaired fashion item images strewn across the digital realm has beckoned for innovative exploration. By leveraging a self-supervised learning approach to explore the stylistic representation of these fashion items, we can ascertain the compatibility rule between individual items, alleviating the prerequisite for meticulously assembled matching outfits datasets, and ultimately empowering the generative synthesis of harmoniously collocated clothing from singular items. Actually, the above-mentioned research problem can be approached as a direct unsupervised image-to-image (I2I) task. The established methods in this task \cite{zhu2017unpaired,huang2018munit,DRIT,DRIT_plus,yang2023gp} aim to learn a mapping function that takes an image from the source domain as input and produces a correspondent image in the target domain without paired data. The core of their strategy is the disentanglement of style and content codes within the images of both domains, thereby enabling the collocation between the input and output during the I2I translation process through self-supervised mechanisms. However, the direct application of these methods to the task of unsupervised CCS may be ineffective. The primary reasons include: (i) in comparison to the default setting of spatial alignment between the input and output of image translation, there is an absence of local spatial alignment relationship between the compatible upper and lower clothing. This is primarily because the matching pair of upper and lower clothing items share harmonious elements, such as stripes and patterns, as opposed to spatial alignment; and (ii) in the learning of fashion compatibility rules, relying solely on disentanglement strategies with self-supervised reconstruction may engender challenges.

In order to address the above drawbacks, we propose a novel approach termed the style- and texture-guided generative network (ST-Net), designed to accomplish the task of unsupervised CCS. In particular, ST-Net employs GAN inversion \cite{abdal2019image2stylegan,richardson2021encoding} as 
its fundamental framework for task execution. Within this framework, an input image of clothing is encoded by ST-Net to obtain a global vector preserving the stylistic representation and overlooking the spatial information of the input. Furthermore, we introduce a self-supervised module dedicated to fashion style and texture prediction, ensuring collocation between model input and output. Moreover, during the training process of ST-Net, we observed challenges in maintaining the visual authenticity of the synthesized images when relying solely on a conventional real/fake discriminator. To augment the supervisory capacity of the generated images, we employ a dual discriminator to enhance the visual fidelity from two distinct perspectives. In order to evaluate the performance of our proposed ST-Net, we constructed a large-scale dataset comprised of upper and lower clothing items. 

The contributions of the paper can be summarized as follows. In this work, we raise a new problem of CCS without the necessity of matching outfits. 
We tackle this issue utilizing a self-driven framework, ST-Net, fortified by supervision derived from the styles and textures of clothing.
For unsupervised CCS, we introduce a style- and texture-guided discriminator to extract the stylistic representation of clothing. It can supervise the ST-Net to generate compatible clothing. Moreover, a dual discriminator is introduced to enhance the visual fidelity of synthesized images
. Extensive experiments substantiate that ST-Net surpasses state-of-the-art benchmarks in terms of visual authenticity and fashion compatibility.

\vspace{-0.3cm}

\section{Methodology}
\vspace{-0.3cm}
\begin{figure}[t]
\centering
\includegraphics[width=0.5 \textwidth]{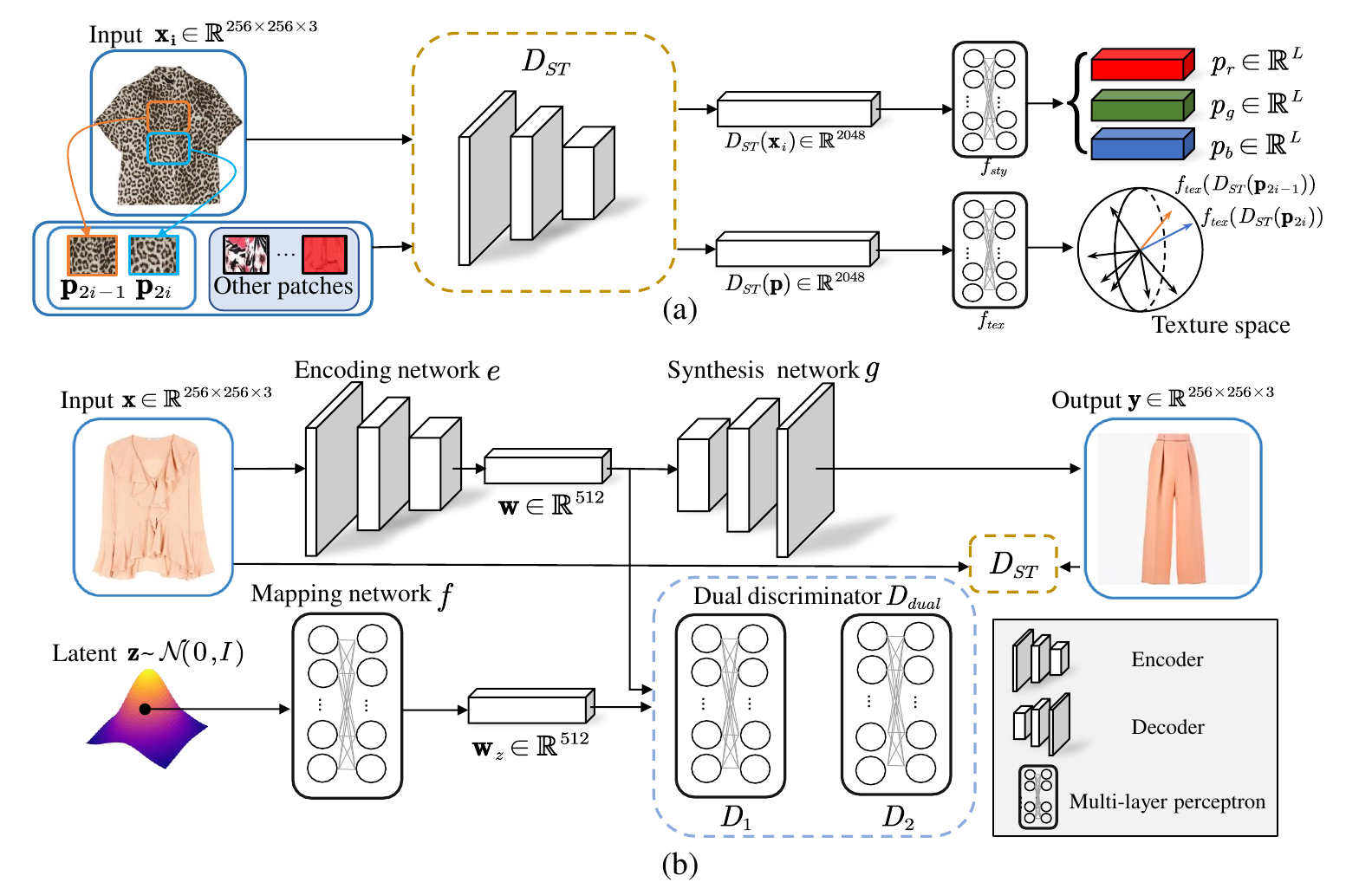}
\vspace{-10mm}
\caption{Overview of our proposed ST-Net: (a) the training phase of $D_{ST}$ and (b) the training phase of the generator $G$.}
\vspace{-0.5cm}
\label{fig1:env}
\end{figure}


\subsection{Proposed Framework}
\vspace{-0.2cm}
Let $\mathcal{X}$ and $\mathcal{Y}$ represent the source and target domains respectively, within the context of the collocated clothing synthesis task. For instance, $\mathcal{X}$ and $\mathcal{Y}$ pertain to the upper and lower clothing domains within the `upper $\rightarrow$ lower' translation direction. Given an image $\mathbf{x}$ from the source domain $\mathcal{X}$, our goal is to train a generator $G$ that can generate an image $\mathbf{y}$ in the target domain $\mathcal{Y}$ that matches to the image $\mathbf{x}$. Fig. \ref{fig1:env} illustrates an overview of ST-Net, which consists of three modules: a pre-trained style- and texture-guided discriminator ${D}_{ST}$, a generator $G$, and a dual discriminator $D_{dual}$.

\textbf{Style- and texture-guided discriminator.} The discriminator ${D}_{ST}$ is tasked with providing supervision to the generator $G$ concerning fashion compatibility, employing a self-supervised approach. In particular, during the training phase of $G$, the ${D}_{ST}$ was trained and then all its parameters were frozen. It employs a standard residual network (ResNet) \cite{he2016deep} as its backbone, generating a feature in the $\mathbb{R}^{2048}$ space. Prior research \cite{kim2021self} has indicated that style and texture significantly contribute to fashion compatibility. To train the ${D}_{ST}$, we have implemented two groups of multi-layer perceptrons (MLPs) in parallel subsequent to this module, dedicated to the classification of style and texture respectively. On one hand, the color scheme significantly influences the global aesthetic of a fashion style. In our model's style classification, we employ the color histogram distribution as an approximate measure. In particular, we utilize a subsequent MLP, named $f_{sty}$, connected in series to constrain $D_{ST}$ to learn the histogram of a given image's color distribution across the three RGB channels. This process allows us to derive a compact representation of a fashion item's style using $D_{ST}$. As depicted in Fig. \ref{fig1:env}, we execute color classification for each fashion item image $\mathbf{x}$. Ultimately, the MLP dedicated to style classification generates an output of ($3 \times L$) neurons, each corresponding to the color distribution across the three RGB channels. Here, $L$ represents the number of histogram bins. On the other hand, texture serves as a local descriptor, playing a pivotal role in the stylistic representation of fashion items. In our implementation, the images of fashion items were cropped at the patch level. Each of these extracted patches was subsequently processed through $D_{ST}$ and an MLP, named $f_{tex}$, for instance discrimination. The patches from the same image are considered to be in the same category as positive samples, and those from different images are classified as other categories as negative samples. By using the aforementioned self-supervision techniques on style and texture, the $D_{ST}$ is capable of learning a compact and efficient representation of fashion items, specifically focusing on their stylistic attributes. 

\textbf{Dual discriminator.} During the training phase of ST-Net, we observed the occurrence of a mode collapse phenomenon. To address this issue and augment the authenticity of the synthesized images, we incorporated a dual discriminator scheme \cite{nguyen2017dual} into our model. In particular, the dual discriminator $D_{dual}$ comprises two sub-discriminators, denoted as $D_1$ and $D_2$, which favor positive and negative samples respectively. As shown in Fig. \ref{fig1:env}, given an image $\mathbf{x}$, the encoding network $e$ initially translates it into a latent space code $\mathbf{w} \in \mathbb{R}^{512}$. Concurrently, a noise $\mathbf{z}$, sampled from a standard Gaussian distribution $\mathcal{N}(0,I)$, is fed into the mapping network $f$ as $\mathbf{w}_{z}=f(\mathbf{z})$, which resides in the $\mathcal{W}$ space. Following this, the discriminator $D_1$ assigns a low score to the latent code $\mathbf{w}$, while attributing a high score to the latent code $\mathbf{w}_{z}$. In contrast, discriminator $D_2$ assigns a high score to $\mathbf{w}$ and a low score to $\mathbf{w}_{z}$. These two discriminators supervise the generator $G$ by minimizing both the Kullback-Leibler (KL) and reverse KL divergences between the model and data distributions. This strategy assists the model in assigning a balanced distribution of probability mass of the generated data distribution, to enhance the overall model performance.

\textbf{Generator.} Given an image $\mathbf{x}$ from the source domain $\mathcal{X}$, $G$ aims to synthesize an item $\mathbf{y}$ in the target domain $\mathcal{Y}$, where the given $\mathbf{x}$ and synthesized $\mathbf{y}$ are expected to be compatible. In particular, the generator $G$ is comprised of a mapping network $f$, an encoding network $e$, and a synthesis network $g$, shown in Fig. \ref{fig1:env}. It should be noted here that $G$ is built upon GAN inversion. A StyleGAN model is pre-trained in each domain of upper and lower in advance to obtain the $f$ and $g$. In the subsequent training phase, the parameters of $f$ and $g$ are frozen. The architecture of the encoding network follows the work of \cite{richardson2021encoding}. Given an image $\mathbf{x}$, the generator $G$ can produce the targeted image $\mathbf{y}$ with $g(e(\mathbf{x}))$.
\vspace{-0.3cm}
\subsection{Training Losses}
\vspace{-0.2cm}
\textbf{Style- and texture-guided discriminator loss.} To train the style- and texture-guided discriminator $D_{ST}$, a style loss and a texture loss are utilized here. The style loss is implemented by regressing the color histogram of the provided image across the three channels of the RGB color space. After obtaining the feature from the $D_{ST}$, the histogram prediction results are computed through the MLP, $f_{sty}$, for style classification. Subsequently, a style loss is implemented on the color histogram prediction results, wherein $p_r$, $p_g$, and $p_b$ mean the respective prediction results for the red, green, and blue channels.
\begin{align}
    \mathcal{L}_{sty} = {D}_{KL}[ p_r || h_{r}] + {D}_{KL}[p_g || h_{g}] + {D}_{KL}[p_b || h_{b}]
    \label{ColorLoos} ,
\end{align}
where $D_{KL}$ is the KL divergence, and $h_{r}$, $h_{g}$, $h_{b}$ are the ground truth RGB histograms obtained from the input image. In the context of texture loss, given a batch of images denoted as $\{\mathbf{x}_i\}^{n}_{i=1}$, we perform a random cropping procedure on each image $\mathbf{x}_i$. This operation results in the creation of two distinct patches, named $\mathbf{p}_{2i-1}$ and $\mathbf{p}_{2i}$. The texture loss is implemented in instance discrimination form as follows:
\begin{align}
    \mathcal{L}_{tex} = -\log{\frac{\exp{(f_{tex}({D}_{ST}(\mathbf{p}_{2i}))^Tf_{tex}({D}_{ST}(\mathbf{p}_{2i-1})))}}
    {\sum_{j=1}^{2\times n} \exp{(f_{tex}({D}_{ST}(\mathbf{p}_{j}))^Tf_{tex}({D}_{ST}(\mathbf{p}_{2i-1})))}}},
    \label{PDLoss}
\end{align}
where $\mathbf{p}_j$ is sampled from all patches in the same batch.

To train the $D_{ST}$, a full objective is provided as follows:
\begin{align}
    \mathcal{L}_{ST}^{train} = \lambda_{sty}\mathcal{L}_{sty}+ \lambda_{tex}\mathcal{L}_{tex},
    \label{STDiscriminatorTrainLoss}
\end{align}
where $\lambda_{sty}$ and $\lambda_{tex}$ are hyper-parameters that are used to balance the relative contributions of the loss terms.

Through training with the aforementioned loss objective, the $D_{ST}$ is capable of extracting a compact style representation based on a clothing image. In particular, given an input image $\mathbf{x}$ and a corresponding output image $\mathbf{y}$ synthesized by the generator $G$, the style- and texture-guided discriminator $D_{ST}$ can yield the subsequent loss:
\begin{align}
    \mathcal{L}_{ST} = 1 - cos(D_{ST}(\mathbf{x}), D_{ST}(\mathbf{y})),
    \label{STLoss}
\end{align}
where $cos$ is a cosine similarity operation. This loss is used for the assurance of an appropriate collocation between the images of the given and the synthesized fashion items.

\textbf{Dual discriminator loss.}
To guarantee the generator $G$ can produce photo-realistic images of clothing, a dual discriminator is incorporated within the ST-Net. The dual discriminator $D_{dual}$, represented by $D_{1}$ and $D_{2}$, along with the encoding network $e$, engages in a three-player minimax optimization game as follows:

\vspace{-0.3cm}
\begin{equation}
\begin{split}
    \min_{G}\max_{D_1, D_2} \mathbb{E}_{\mathbf{z}\sim {\mathcal{N}}}[\log{D_1(f(\mathbf{z})))}] + 
    \mathbb{E}_{\mathbf{x}\sim \mathcal{X}}
    [\log{D_2(e(\mathbf{x}))}]  
    + \\
    \mathbb{E}_{\mathbf{x}\sim {\mathcal{X}}}[1 -  \log{D_1(e(\mathbf{x})))}]
    + \mathbb{E}_{\mathbf{z}\sim \mathcal{N}}[1 - \log{D_2(f(\mathbf{z}))}].
\end{split}
\label{DualGanloss}
\end{equation}



\textbf{Total loss.} When training the generator $G$ of ST-Net, the total loss function can be summarized as follows:
\begin{equation}
\begin{split}
    \mathcal{L}_{total} & = \lambda_{ST} \mathcal{L}_{ST} + 
    \mathbb{E}_{\mathbf{x}\sim {\mathcal{X}}}[1 -  \log{D_1(e(\mathbf{x}))}] \\ & +
    \mathbb{E}_{\mathbf{x}\sim \mathcal{X}}
    [\log{D_2(e(\mathbf{x}))}], 
\end{split}
\label{encoderLoss}
\end{equation}
where $\lambda_{ST}$ is the hyper-parameter that weights the $\mathcal{L}_{ST}$.
\vspace{-0.3cm}

\section{Experiments}
\vspace{-0.3cm}
\subsection{Implemention Details}
\vspace{-0.2cm}

\textbf{Dataset.} Given that the datasets collected for CCS in previous studies \cite{liu2019collocating,yu2019personalized,liu2019toward} are neither publicly accessible nor available, we have constructed a large-scale dataset to facilitate the task of unsupervised CCS. Our dataset encompasses 30, 876 images, covering both upper and lower clothing items. We partition the dataset into training and test sets, maintaining a ratio of $4:1$, for both `upper $\rightarrow$ lower' and `lower $\rightarrow$ upper'  settings. Within this dataset, there exist instances where certain fashion items recur. To ensure non-overlap between the training and test sets, we employ a graph-based method and similarity comparison with learned perceptual image patch similarity (LPIPS) \cite{zhang2018unreasonable}. This ensures that the source domain images in any transformation direction do not intersect. For instance, in the transition direction of `upper $\rightarrow$ lower', there is no intersection between the upper clothing images in the training set and those in the test set. 
All images in the dataset are with a resolution of $256\times 256$ pixels. The dataset can be accessed at \url{https://huggingface.co/datasets/Fashion-Intelligence/ST-Net-dataset}.

\textbf{Network training.} In our implementation, the batch size for training ST-Net was set to 32. All experiments were performed on a single A6000 graphics card, and the implementation was carried out in PyTorch. ST-Net was optimized with the Adam optimizer with $\beta_1 = 0$ and $\beta_2 = 0.99$, and its learning rate was set to $2 \times 10^{-4}$. 
$L$ was set to ten. 
$\lambda_{sty}$, $\lambda_{tex}$, and $\lambda_{ST}$ were set to 1, 2.2, and 1, respectively.

\vspace{-0.5cm}

\begin{figure}
\centering
\includegraphics[width=0.436 \textwidth]{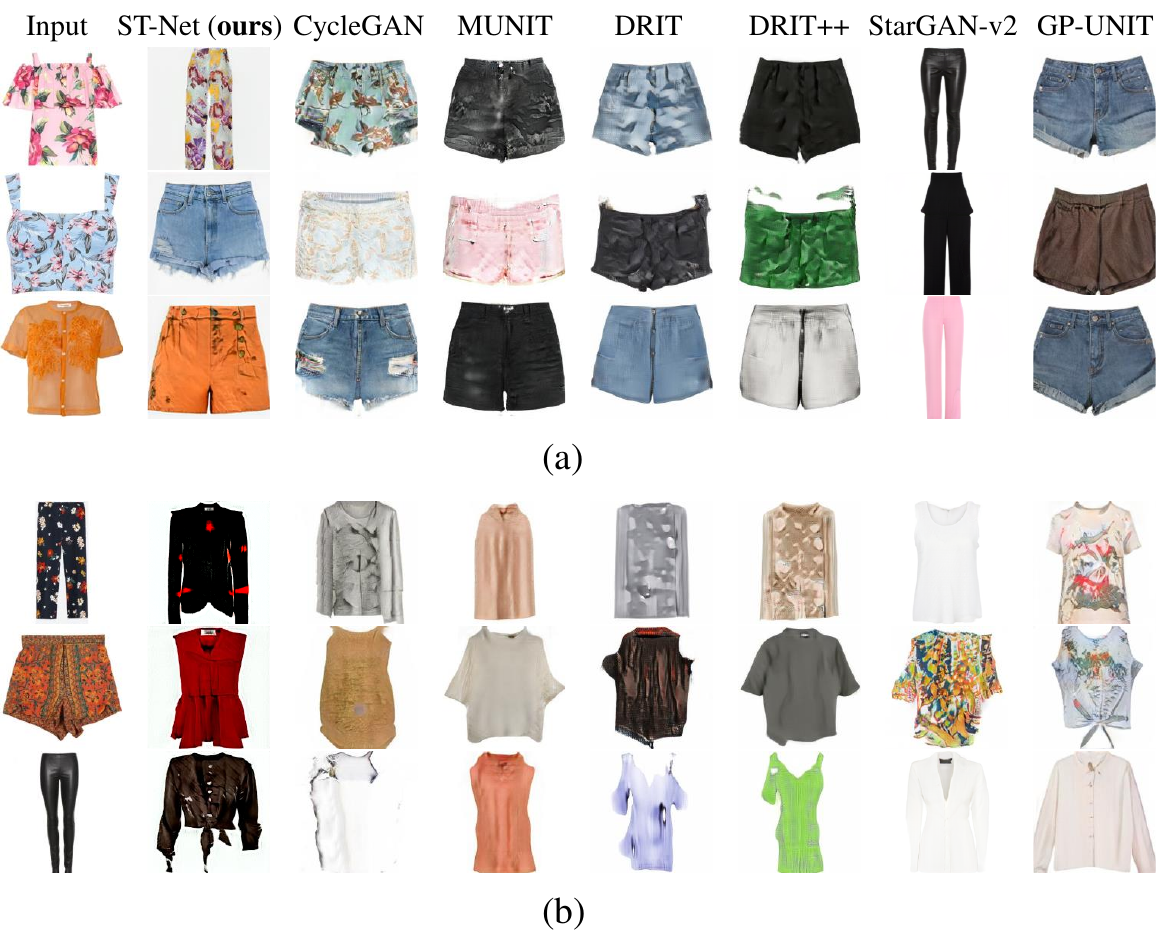}
\vspace{-6mm}
\caption{Comparison between ST-Net and other baselines, which are CycleGAN \cite{zhu2017unpaired}, MUNIT \cite{huang2018munit}, DRIT \cite{DRIT}, DRIT++ \cite{DRIT_plus}, StarGAN-v2 \cite{choi2020stargan}, and GP-UNIT \cite{yang2023gp}: (a) `upper $\rightarrow $ lower' direction and (b) `lower $\rightarrow $ upper' direction. 
}
\label{fig2:comparison}
\vspace{-0.6cm}
\end{figure}

\begin{table}[t]
\vspace{-1.2mm}
  \centering
  \small
  \caption{Comparative analysis of ST-Net, its variants, and other baselines (here, `GP-UNIT w/ $\mathcal{L}_{ST}$' denotes the GP-UNIT model augmented with $\mathcal{L}_{ST}$; `ST-Net w/o $\mathcal{L}_{ST}$' and `ST-Net w/o $\mathcal{L}_{dual}$' represent the variants of ST-Net without the $D_{ST}$ and dual discriminator, respectively)}
  \label{tab:cmp}
  \setlength{\tabcolsep}{1.5mm}{
  \begin{tabular}{lcc|cc}
  \toprule[1pt]
    \multirow{2}{*}{Method} & \multicolumn{2}{c}{upper $\rightarrow $ lower} & \multicolumn{2}{c}{lower $\rightarrow $ upper} \\ \cline{2-5}
    & FID ($\downarrow$) & FCTS ($\uparrow$) & FID ($\downarrow$) & FCTS ($\uparrow$) \\ 
    \hline
    CycleGAN \cite{zhu2017unpaired}& 30.607 & 0.650 & 98.810 & 0.593\\
    MUNIT \cite{huang2018munit} & 40.314 & 0.508 & 62.079 & 0.523\\
    DRIT \cite{DRIT} & 88.485 & 0.504 & 62.660 & 0.506\\
    DRIT++ \cite{DRIT_plus} & 81.401 & 0.508 & 45.322 & 0.513\\
    StarGAN-v2 \cite{choi2020stargan}& 98.817 & 0.510 & 49.191 & 0.553\\
    GP-UNIT \cite{yang2023gp} & 43.539 & 0.502 & 43.740 & 0.497\\
    \hline
    GP-UNIT w/ $\mathcal{L}_{ST}$ & 39.903 & 0.551 & 44.529 & 0.538\\
    \hline
    ST-Net w/o $\mathcal{L}_{ST}$ & 27.128 & 0.477 & \textbf{41.887} & 0.477\\
    ST-Net w/o $\mathcal{L}_{dual}$ & 28.813 & 0.702 & 52.955 & 0.604\\
    ST-Net (\textbf{ours}) & \textbf{26.659} & \textbf{0.706} & 43.614 & \textbf{0.612}\\
\bottomrule[1pt]
  \end{tabular}}
  \vspace{-0.6cm}
\end{table}

\subsection{Comparison to State-of-the-art Methods}
\vspace{-0.2cm}

\textbf{Evaluation metrics.} To substantiate the efficacy of the proposed method, we evaluate both the authenticity of visually generated images and the degree of match between synthesized and given fashion items. For the first evaluation metric, we employ the Fréchet inception distance (FID) \cite{heusel2017gans}, whereby a lower score signifies that the distribution of generated images more closely mirrors that of real images. As for the second evaluation criterion, we utilize a fashion compatibility test score (FCTS) \cite{zhou2022learning}, where a higher score signifies a higher degree of collocation for a given fashion item.


\textbf{Baselines.} We compare our proposed method with state-of-the-art unsupervised image-to-image translation models, including CycleGAN \cite{zhu2017unpaired}, MUNIT \cite{huang2018munit}, DRIT \cite{DRIT}, DRIT++ \cite{DRIT_plus}, StarGAN-v2 \cite{choi2020stargan}, and GP-UNIT \cite{yang2023gp}. 

\textbf{Qualitative comparison.} We present the synthesized results from ST-Net and other baseline models for CCS, covering both `upper $\rightarrow$ lower' and `lower $\rightarrow$ upper' transformations. As depicted in Fig. \ref{fig2:comparison} (a), it is observable that all baseline models, with the exception of StarGAN-v2, only generate images of lower clothing that bear a resemblance in shape to the provided upper clothing. Additionally, StarGAN-v2 tends to generate lower clothing with elongated legs. In contrast, our method yields the most diverse and photo-realistic results. For the `lower $\rightarrow$ upper' transformation, as shown in Fig. \ref{fig2:comparison} (b), a similar phenomenon is evident. Moreover, in comparison to other baseline models, ST-Net generates images of clothing that exhibit a greater harmony in elements with the provided clothing. These findings indicate that our method outperforms other baselines in terms of both visual authenticity and fashion compatibility.

\textbf{Quantitative comparison.} We performed quantitative analysis, as illustrated in Table \ref{tab:cmp}, primarily concentrating on visual authenticity (FID) and fashion compatibility (FCTS). From the standpoint of visual authenticity, our method has attained the lowest FID score, signifying that the results produced by ST-Net are the most photo-realistic. Concurrently, ST-Net surpasses other baseline models in fashion compatibility, as corroborated by the FCTS metric depicted in Table \ref{tab:cmp}. Taking into account both visual authenticity and fashion compatibility, the proposed ST-Net model not only assures the authenticity of the generated image but also keeps collocation between the given and synthesized clothing images. 

\vspace{-0.5cm}

\subsection{Ablation Study}
\vspace{-0.2cm}
To validate the effectiveness of components in our proposed framework, three sets of experiments were carried out. Firstly, among all the compared baselines, GP-UNIT demonstrates superior performance in terms of visual authenticity. This model, like ours, also utilizes a pre-trained GAN as the backbone. We incorporated the $D_{ST}$ module into this model to further substantiate the efficacy of the $D_{ST}$. Experimental results in Table \ref{tab:cmp} indicate that the inclusion of the $D_{ST}$ enhances the model's performance in terms of fashion matching, while the image authenticity metric remains largely unchanged. Nevertheless, our model still surpasses the performance of GP-UNIT even with the added $D_{ST}$. Secondly, we performed ablation studies on the $D_{ST}$, and the results presented in Table \ref{tab:cmp} validated the effectiveness of $D_{ST}$ in fashion compatibility. Thirdly, we conducted ablation experiments on the $D_{dual}$, and the results, as shown in Table \ref{tab:cmp}, reveal that our dual discriminator plays a pivotal role in the visual fidelity of synthesized images.

\vspace{-0.2cm}

\section{Conclusion}
\vspace{-0.2cm}
This paper has proposed a self-driven framework for collocated clothing synthesis without the dependency on constructed matching outfits, thereby mitigating the laborious data collection burden. Central to our approach is the integration of a fashion compatibility supervision module, which is trained with the assistance of a style- and texture-guided discriminator, leveraging the principles of self-supervised learning. Additionally, a dual discriminator is implemented to augment the quality of the synthesized clothing images. Experimental results confirmed the superiority of our method compared with other baselines. 
In our future work, we envision a continued exploration of more potent information sources to refine the generation of collocated clothing.

\vfill\pagebreak



\bibliographystyle{IEEEbib}
\bibliography{refs}

\end{document}